\documentclass[11pt]{article}

\usepackage[preprint]{acl}

\usepackage{times}

\usepackage{latexsym}

\usepackage[T1]{fontenc}

\usepackage[utf8]{inputenc}

\usepackage{microtype}

\usepackage{inconsolata}

\usepackage{graphicx}
\usepackage{bm}
\usepackage{amsmath}
\usepackage{amssymb}
\usepackage{mathtools}
\usepackage{amsthm}
\usepackage{wrapfig}
\usepackage{url}
\usepackage[table]{xcolor}
\usepackage{float}
\usepackage{microtype}
\usepackage{graphicx}
\usepackage{subcaption}
\usepackage{booktabs}
\usepackage{amsmath}
\usepackage{bbm}
\usepackage{pifont}
\usepackage{multirow}
\usepackage{multicol}
\usepackage{makecell}
\usepackage{amssymb}
\usepackage{xspace}
\usepackage{tcolorbox}
\definecolor{ForestGreen}{RGB}{34,139,34}
\definecolor{BrickRed}{rgb}{.72,0,0}
\definecolor{LakeBlue}{RGB}{0,61,153}
\definecolor{darkgreen}{rgb}{0.0, 0.5, 0.0}
\definecolor{MiOrange}{RGB}{255,225,204}
\definecolor{TableBlue}{RGB}{218,236,252}
\definecolor{PromptBlue}{RGB}{154,198,231}
\usepackage{paralist}
\usepackage{booktabs}
\hypersetup{citecolor=LakeBlue}
\usepackage{tabularx} 
\usepackage[T1]{fontenc}
\usepackage[utf8]{inputenc}

\title{Enhancing Trustworthy GUI Grounding via Self-Critiqued \\ Reinforcement Learning}

\author{
    \makecell[l]{Shaojie~Zhang$^*$,~
                Pei~Fu$^*$,~
                Ruoceng~Zhang,~
                Jiahui~Yang,~
                Anan~Du,~
                Xiuwen~Xi,~ \\
                Shaokang~Wang,~ 
                Ying~Huang,~ 
                Bin~Qin,~
                Zhenbo~Luo$^\dagger$,~
                Jian~Luan} \\
    ~~ \\
MiLM Plus, Xiaomi Inc\\
\texttt{\{zhangshaojie5, fupei1, luozhenbo, luanjian\}@xiaomi.com}
\vspace{-6pt} \\
}

\begin{document}
\maketitle
\begin{abstract}
Autonomous graphical user interface (GUI) agents rely on accurate GUI grounding, which maps language instructions to on-screen coordinates, to execute user commands. However, current models, whether trained via supervised fine-tuning (SFT) or reinforcement learning (RL), often provide confidence signals that are poorly aligned with actual grounding correctness, leading to overconfident and unreliable predictions. To address this, we propose HyperClick, a novel framework that enhances trustworthy GUI grounding through self-critiqued reinforcement learning (SCRL). HyperClick combines a correctness reward and a confidence alignment reward, training the policy model to output both a click prediction and an explicit confidence estimate. This approach jointly optimizes grounding accuracy and confidence reliability through confidence-based self-assessment. Extensive experiments on challenging benchmarks show that HyperClick maintains strong grounding performance while providing better-aligned confidence estimates. By exposing uncertainty alongside GUI actions, HyperClick supports confidence-based abstention in GUI automation. Code will be released \href{https://github.com/xiaomi-research/hyperclick}{here}.
\end{abstract}

\begin{figure*}[!htbp]
    \centering
    \includegraphics[width=0.98\linewidth]{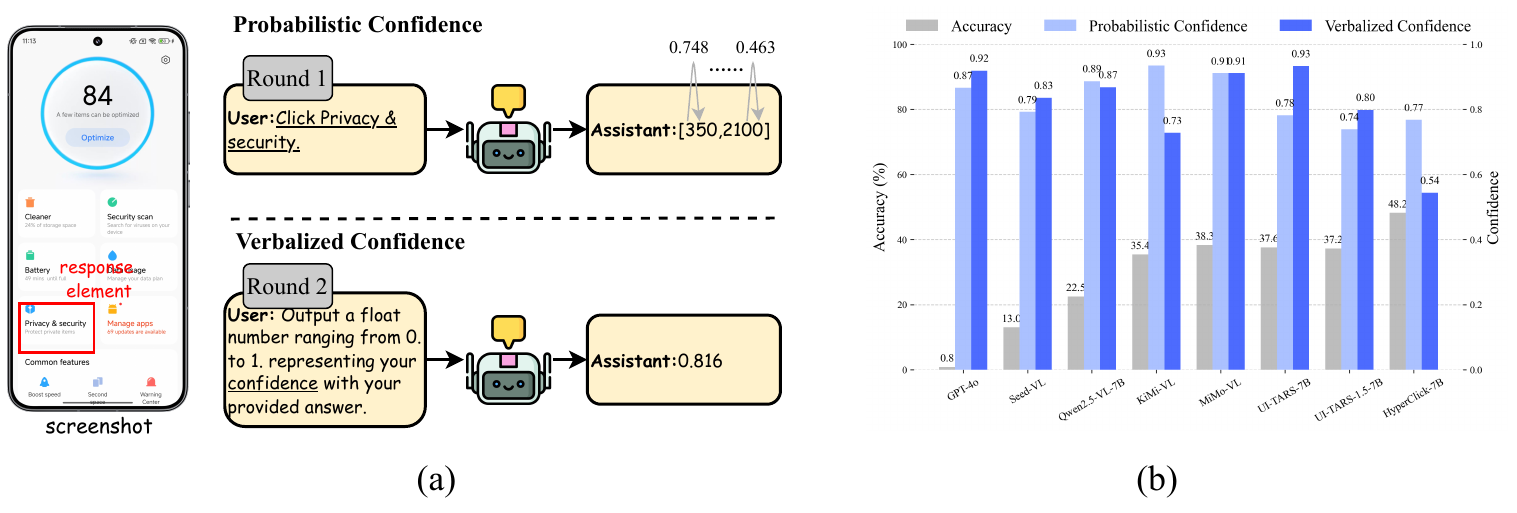}
    \caption{Overview of accuracy and confidence evaluation on ScreenSpot-Pro. (a): Illustration of probabilistic and verbalized confidence. Probabilistic confidence represents the probability of the model generating the next token corresponding to the target coordinates, while verbalized confidence indicates the model’s self-reported certainty about its output in natural language. (b): Comparisons of accuracy, probabilistic confidence, and verbalized confidence for several general-purpose and GUI-specific models on the ScreenSpot-Pro benchmark. The models exhibit a higher confidence in their answers than in the accuracy that they actually achieve.}
    \label{fig: motivation}
\end{figure*}

\section{Introduction}

The revolution of autonomous graphical user interface (GUI) agents is transforming human-computer interaction, allowing users to control mobile applications, web platforms, and complex desktop software directly through natural language instructions~\citep{wang2024gui,nguyen2024gui}. At the heart of these agents lies the GUI grounding, the ability to accurately map textual commands to precise pixel coordinates on user interface elements~\citep{cheng2024seeclick,tang2025gui}. This fundamental task determines whether an agent can successfully execute user commands, making it the cornerstone of the GUI automation.

Recent progress in GUI grounding has been driven by supervised fine-tuning (SFT) with curated large-scale datasets~\citep{wu2024atlas,gou2025uground,xu2024aguvis} and reinforcement learning (RL) with verifiable GUI-specific rewards~\citep{lu2025ui,luo2025gui,liu2025infigui}. Although these techniques yield strong performance, their confidence estimates are often poorly aligned with actual grounding correctness, making it difficult to judge when predictions are reliable.

\textbf{A trustworthy GUI agent should be aware of its limitations and distinguish accurately between what it can and cannot do}~\citep{ding2025lvlms}. Although confidence estimation has been extensively studied in large language models (LLMs)~\citep{xiong2023can,tian2023just}, it remains underexplored in GUI agents. The reliability level of an agent can be assessed by the alignment between its confidence and actual performance~\citep{ding2025lvlms}. In this paper, we first evaluate probabilistic and verbalized confidence for several general models~\citep{openai2024hello,bai2025qwen2,guo2025seed1,team2025kimi,coreteam2025mimovltechnicalreport} and GUI-specific models~\citep{qin2025ui} on the ScreenSpot-Pro benchmark~\citep{li2025screenspot}, which emphasizes high-resolution displays, smaller target sizes, and complex environments. Specifically, probabilistic confidence reflects token-level likelihoods for predicted coordinates~\citep{guo2017calibration,desai2020calibration}, while verbalized confidence captures self-reported certainty in natural language~\citep{lin2022teaching,yang2024alignment}.

As shown in Figure~\ref{fig: motivation}, the models exhibit a higher confidence in their answers than in the accuracy that they actually achieve. In other words, even on challenging tasks, these agents remain overconfident in their predictions, both probabilistically and in their self-assessments. This resembles the broader reliability challenge commonly observed in LLMs and vision-language models (VLMs), where models can produce erroneous outputs while maintaining high confidence~\citep{ji2023survey,ji2023towards,kalai2025language}. This limitation is particularly critical in real-world GUI tasks, where their dynamic, continuous nature means that even a single error at an intermediate step can result in overall task failure.

To address this limitation, we propose HyperClick, a novel framework that enhances trustworthy GUI grounding through self-critiqued reinforcement learning (SCRL). Unlike prior approaches that treat grounding as a binary, hit-or-miss classification task, HyperClick explicitly integrates confidence estimation into the decision-making process. Specifically, each prediction delivers not only a targeted UI element but also a verbalized confidence statement that serves as a reliable self-assessment. In this work, we deliberately focus on aligning verbalized confidence rather than probabilistic confidence. While the latter relies heavily on token likelihoods and is easily confounded by coordinate tokenization or numeric formatting, verbalized confidence is entirely model-agnostic and can be directly exposed to downstream GUI agents, making it a cleaner reflection of spatial click quality.

Specifically, we introduce two complementary rule-based reward mechanisms that optimize both action precision and confidence alignment. A binary reward enforces correct grounding positions, while a confidence alignment reward aligns the verbalized confidence of policy models with a bounded spatial confidence target constructed over the annotated element region. This dual mechanism enables HyperClick to achieve two intertwined goals: accurate GUI grounding and better-aligned confidence. By training the model to expose uncertainty alongside its click prediction, HyperClick reduces high-confidence errors and enables confidence-based abstention for safer GUI decision-making.

Our contributions are summarized as follows:
\begin{itemize}
    \item We systematically reveal that existing GUI grounding models are prone to overconfident predictions and highlight their implications for reliable GUI automation.
    \item We propose HyperClick, a trustworthy GUI grounding framework that integrates SCRL, introducing a dual reward mechanism that jointly optimizes grounding accuracy and confidence alignment via binary correctness and a bounded spatial confidence target.
    \item Through extensive evaluations on challenging GUI grounding benchmarks, HyperClick maintains strong accuracy while producing confidence-quality-aligned estimates that enable selective abstention on uncertain clicks.
\end{itemize}

\section{Related Work}

\subsection{GUI Agents and GUI Grounding}
GUI agents automate desktop and mobile tasks by interacting with graphical user interfaces through natural language instructions~\citep{wang2024gui,nguyen2024gui,zhang2024large}. Recent VLM-based agents~\citep{cheng2024seeclick,wu2024atlas,qin2025ui} combine visual perception with language reasoning to handle diverse interface styles. Their reliability largely depends on GUI grounding, which maps instructions to precise interface elements or pixel coordinates.

Early studies mainly acquire GUI-specific abilities through supervised fine-tuning on large-scale GUI corpora~\citep{cheng2024seeclick, lin2025showui, yang2024aria}. SeeClick~\citep{cheng2024seeclick} introduces visual-only GUI grounding, while OS-Atlas~\citep{wu2024atlas}, UGround~\citep{gou2025uground}, and Aguvis~\citep{xu2024aguvis} improve perception by fine-tuning pretrained models on diverse GUI environments. UI-TARS~\citep{qin2025ui} further scales this direction toward an end-to-end GUI agent for unified cross-platform action modeling.

Recent RL work further extends GUI grounding with verifiable rewards inspired by reasoning models~\citep{guo2025deepseek}. R1-style GUI agents optimize policies with such rewards~\citep{lu2025ui,luo2025gui,liu2025infigui,zhang2025btl}, often requiring explicit reasoning before prediction. Later methods tailor reward design to spatial grounding through controllable box-size rewards~\citep{zhou2025gui}, self-evolution with continuous rewards~\citep{yuan2025enhancing}, and Gaussian reward modeling~\citep{tang2025gui}. These methods improve grounding accuracy, but they primarily optimize localization itself and largely overlook whether model confidence is aligned with grounding quality.

\subsection{Confidence in LLMs}
Confidence quantifies prediction reliability and has been widely used in error analysis~\citep{oberkampf2002error} and computer vision tasks such as object detection and segmentation~\citep{ren2015faster,redmon2016you,long2015fully,he2017mask}. For LLMs and VLMs, existing confidence signals can be broadly grouped into \textbf{probabilistic confidence} from token probabilities~\citep{guo2017calibration,desai2020calibration}, \textbf{answer consistency confidence} from agreement among multiple outputs~\citep{zhang2023sac3,manakul2023selfcheckgpt,fu2025deep}, and \textbf{verbalized confidence} from explicit self-reported certainty~\citep{lin2022teaching,yang2024alignment}.

Probabilistic and consistency-based confidence often correlate better with model performance, but they require token likelihoods, repeated sampling, or in-domain calibration. Verbalized confidence is easier to expose to downstream agents and is model-agnostic, yet it is prone to weak alignment and overconfidence. Therefore, we align verbalized confidence with a spatial confidence distribution over the screenshot, turning self-critiqued confidence into a grounding-quality signal for selective abstention.

\section{Method}

\begin{figure*}[htbp]
    \centering
    \includegraphics[width=0.98\linewidth]{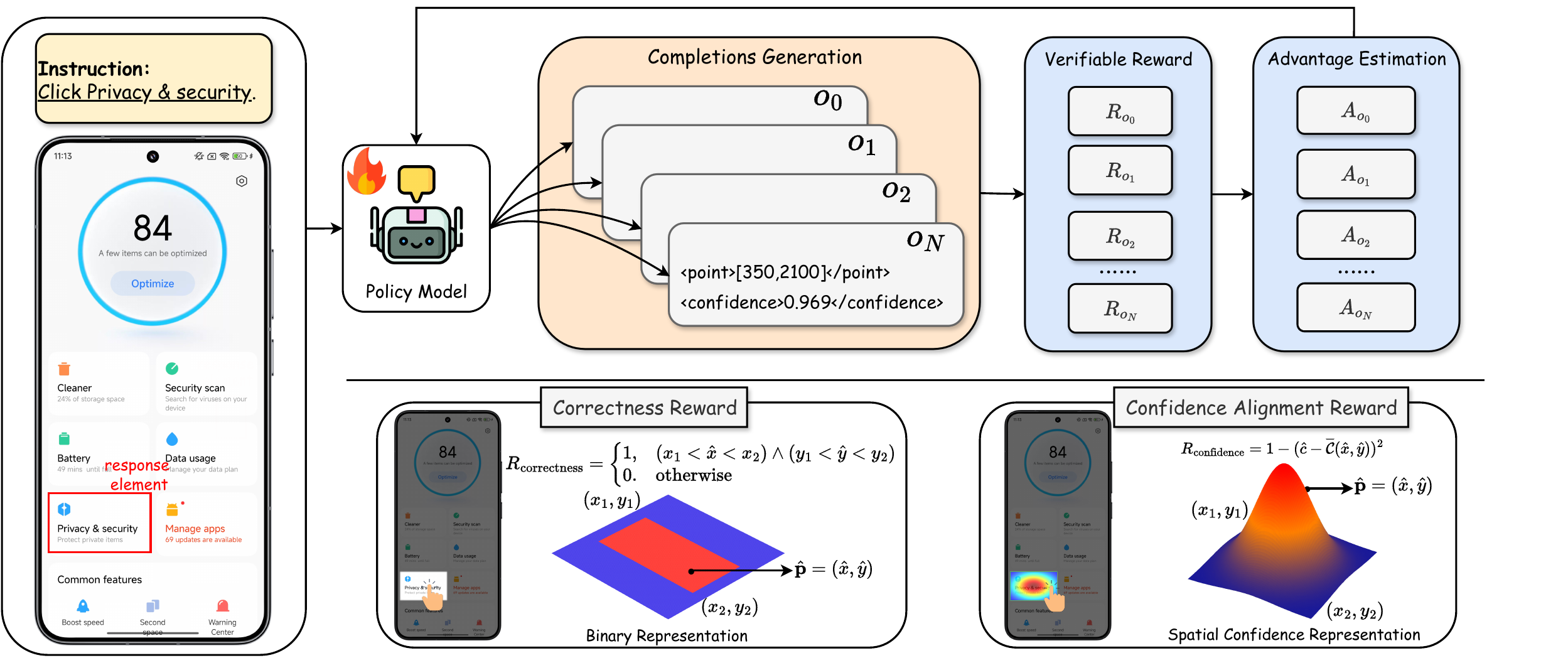}
    \caption{Framework of the proposed HyperClick, optimized with Group Relative Policy Optimization (GRPO). Given a screenshot and an instruction, the policy generates $N$ predictions, which are evaluated by a verifiable reward mechanism. The correctness reward measures grounding precision, while the confidence alignment reward aligns the verbalized confidence of the policy model with a bounded spatial confidence target over the annotated element region. For clarity, the reference model is omitted.}
    \label{fig: Framework}
\end{figure*}

\subsection{Problem Formulation}
GUI grounding maps a natural language instruction to the spatial coordinates of the target UI element. From a policy optimization view, it is commonly instantiated as either location formulation~\citep{wu2024atlas,tang2025gui} or click formulation~\citep{xu2024aguvis,luo2025gui,yuan2025enhancing}.

\begin{itemize}
    \item Location formulation: Given a screenshot $s$ and an instruction $q$, the policy model is optimized by predicting the bounding box $\hat{\mathbf{b}} = (\hat{x}_1, \hat{y}_1, \hat{x}_2, \hat{y}_2)$, where $(\hat{x}_1, \hat{y}_1)$ and $(\hat{x}_2, \hat{y}_2)$ denote the top-left and bottom-right corners of the UI element referred to by $q$.
    \item Click formulation: Alternatively, the policy model predicts a single point $\hat{\mathbf{p}} = (\hat{x}, \hat{y})$, corresponding to the center of the target element, which directly simulates a clicking action.
\end{itemize}

We adopt the click formulation because it directly matches executable GUI actions, reduces the action space, and provides a natural reinforcement learning objective.

\subsection{Confidence Modeling}
\label{confidence_modeling}
Following Gaussian error analysis~\citep{gauss1809theoria,gauss1877theoria,mackenzie1988history}, we construct a bounded spatial confidence representation for GUI grounding using a Gaussian kernel over the annotated element region. This target is not intended to be an absolute probability; instead, it provides a smooth supervision signal for confidence alignment, assigning higher values to clicks near the annotated element center and lower values to less precise clicks. We refer to this objective as \emph{confidence-quality alignment}: confidence that ranks predictions by grounding success and supports selective abstention, rather than an absolute probability over arbitrary failure modes. Since most GUI annotations are bounding boxes, this target naturally combines two requirements: a click outside the target box should receive zero confidence, while clicks inside the box can still be differentiated by their spatial proximity to the center.

For a target box $\mathbf{b}$, let $\bm{\mu}=(\mu_x,\mu_y)$ denote its center and let $\sigma_x,\sigma_y$ control a diagonal covariance. For each predicted point $\hat{\mathbf{p}}=(\hat{x},\hat{y})$, the base spatial confidence target is:
\begin{equation}
    \begin{aligned}
    \mathcal{C}(\hat{\mathbf{p}})
    = \exp\!\left(-\,\tfrac{1}{2}
    \Big[\tfrac{(\hat{x}-\mu_x)^2}{\sigma_x^2}
    + \tfrac{(\hat{y}-\mu_y)^2}{\sigma_y^2}\Big]\right).
    \end{aligned}
\end{equation}

The value is maximized at the ground-truth center. We then constrain the target to the bounding box $\mathbf{b}$ by setting $\overline{\mathcal{C}}(\hat{\mathbf{p}})=\mathcal{C}(\hat{\mathbf{p}})$ when $\hat{\mathbf{p}}\in\mathbf{b}$ and $\overline{\mathcal{C}}(\hat{\mathbf{p}})=0$ otherwise. This preserves the binary correctness boundary used by GUI grounding benchmarks while providing a graded target for confidence learning inside valid regions.
\textbf{Adaptive Variance.} 
Prior work~\citep{zhou2025gui,tang2025gui} shows that smaller UI elements are harder to ground. To handle diverse element sizes, we scale the variance by the box dimensions:
\begin{equation}
    \sigma_x = \alpha \cdot (x_2 - x_1), \quad \sigma_y = \alpha \cdot (y_2 - y_1),
\end{equation}
where $\alpha$ controls how element size affects the standard deviations. As a result, larger elements receive broader high-confidence regions, while smaller elements produce sharper confidence peaks that reflect the higher precision required for successful clicks.

\subsection{Training Objective}
\textbf{Format Reward.}
We constrain the policy to emit each response in the format \texttt{<point>\allowbreak[x,y]\allowbreak</point>\allowbreak <confidence>\allowbreak conf\allowbreak</confidence>}, where \texttt{conf} is rounded to three decimal places. $R_{\text{format}}$ returns $1$ when the completion matches this template and $0$ otherwise, preventing overly long completions and enabling unambiguous parsing.

\textbf{Correctness Reward.}
We use a binary reward to guide the predicted point $\hat{\mathbf{p}}$ into the bounding box $\mathbf{b}$, directly matching the success criterion of GUI grounding:
\begin{equation}
\begin{aligned}
R_{\text{correctness}}
&= \mathbbm{1}_{\hat{\mathbf{p}} \in \mathbf{b}} =
\begin{cases}
    1,  &
    \begin{aligned}[t]
    &\text{if } (x_1 < \hat{x} < x_2) \\
    &\land\, (y_1 < \hat{y} < y_2),
    \end{aligned}\\[4pt]
    0,  & \text{otherwise}.
\end{cases}
\end{aligned}
\end{equation}

\textbf{Confidence Alignment Reward.} 
The confidence alignment reward encourages the policy to evaluate its own prediction. We align the generated confidence $\hat{c}$ with the spatial confidence target in Section~\ref{confidence_modeling} using an $l_2$ constraint:
\begin{equation}
    R_{\text{confidence}} = 1 - (\hat{c} - \overline{\mathcal{C}}(\hat{x}, \hat{y}))^2 .
\end{equation}
The reward is bounded because both $\hat{c}$ and $\overline{\mathcal{C}}(\cdot)$ lie in $[0,1]$, and it favors confidence values close to the constructed target. It also rewards low confidence on incorrect predictions, encouraging confidence-quality alignment rather than overconfident clicks.

The final reward combines format, correctness, and confidence alignment terms:
\begin{equation}
    R = R_{\text{format}} + R_{\text{correctness}} + R_{\text{confidence}}.
\end{equation}
We optimize HyperClick with Group Relative Policy Optimization (GRPO)~\citep{shao2024deepseekmath}. Given $N$ generations $\{o_i\}_{i=1}^N$, GRPO evaluates each output with $R$ and normalizes rewards within the group to obtain relative advantages:
\begin{equation}
    A  _i = \frac{R(o_i) - \text{mean}(\{R(o_j)\}_{j=1}^N)}{\text{std}(\{R(o_j)\}_{j=1}^N)}
\end{equation}
The training objective of GRPO is then defined as 
\begin{equation}
\begin{aligned}
    \mathcal{J}(\theta)
    &= \mathbb{E}_{\{o_i\}\sim \pi_{\theta_{\text{old}}}(\cdot|s,q)}
    \Bigg[
    \frac{1}{N}\sum_{i=1}^{N}
    \min\Big(
    r_i(\theta)A_i,\\
    &\qquad
    \text{clip}(r_i(\theta),1-\epsilon,1+\epsilon)\,A_i
    \Big)
    \Bigg] \\
    &\quad
    -\,\beta\,\text{KL}\!\left(\pi_{\theta}\,\middle\|\,\pi_{\text{ref}}\right),
\end{aligned}
\end{equation}

where $\pi_{\theta}$ is the policy model, $\epsilon$ controls clipping, and $\beta$ weights the KL regularization~\citep{schulman2017proximal,shao2024deepseekmath}.

\section{Experiments}
\label{sec: experiments}
\begin{table*}[!htbp]
\centering
\caption{GUI grounding accuracy on the ScreenSpot~\citep{cheng2024seeclick} and ScreenSpot-V2~\citep{cheng2024seeclick} benchmarks over the Mobile, Desktop, and Web sub-tasks. \textbf{Bold} and \underline{underline} indicate the best and second-best results.}
\resizebox{\linewidth}{!}
{
    \begin{tabular}{lccccccccccccccc}
        \toprule
        \multirow{3}{*}{\textbf{Model}} & \multirow{3}{*}{\textbf{Size}} & \multicolumn{6}{c}{\textbf{ScreenSpot}} & \multirow{3}{*}{\makecell{\textbf{SS} \\ \textbf{Avg.}}} & \multicolumn{6}{c}{\textbf{ScreenSpot V2}} & \multirow{3}{*}{\makecell{\textbf{SSv2} \\ \textbf{Avg.}}} \\
        \cmidrule(lr){3-8}
        \cmidrule(lr){10-15}
        & & \multicolumn{2}{c}{\textbf{Mobile}} & \multicolumn{2}{c}{\textbf{Desktop}} & \multicolumn{2}{c}{\textbf{Web}} & & \multicolumn{2}{c}{\textbf{Mobile}} & \multicolumn{2}{c}{\textbf{Desktop}} & \multicolumn{2}{c}{\textbf{Web}} \\
        \cmidrule(lr){3-4}
        \cmidrule(lr){5-6}
        \cmidrule(lr){7-8}
        \cmidrule(lr){10-11}
        \cmidrule(lr){12-13}
        \cmidrule(lr){14-15}
        & & Text & Icon & Text & Icon & Text & Icon & & Text & Icon & Text & Icon & Text & Icon \\
        \midrule
        \multicolumn{16}{l}{\cellcolor{TableBlue}{\textit{General Models}}} \\
        GPT-4o~\citep{openai2024hello} 
        & - & 30.5 & 23.2 & 20.6 & 19.4 & 11.1 & 7.8 & 18.8 & 26.6 & 24.2 & 24.2 & 19.3 & 12.8 & 11.8 & 20.1 \\
        Qwen2.5-VL~\citep{bai2025qwen2}
        & 7B & - & - & - & - & - & - & 84.7 & 97.6 & 87.2 & 90.2 & 74.2 & 93.2 & 81.3 & 88.8\\
        \midrule
        \multicolumn{16}{l}{\cellcolor{TableBlue}{\textit{GUI-specific Models (SFT)}}} \\
        SeeClick~\citep{cheng2024seeclick} 
        & 9.6B & 78.0 & 52.0 & 72.2 & 30.0 & 55.7 & 32.5 & 53.4 & 78.4 & 50.7 & 70.1 & 29.3 & 55.2 & 32.5 & 55.1\\
        UGround~\citep{gou2025uground} 
        & 7B & 82.8 & 60.3 & 82.5 & 63.6 & 80.4 & 70.4 & 73.3 & 75.1 & 84.5 & 85.1 & 61.4 & 84.6 & 71.9 & 76.3\\
        OS-Atlas~\citep{wu2024atlas}
        & 7B & 93.0 & 72.9 & 91.8 & 62.9 & 90.89 & 74.3 & 82.5 & 95.2 & 75.8 & 90.7 & 63.6 & 90.6 & 77.3 & 84.1 \\
        UI-TARS~\citep{qin2025ui}
        & 7B & 94.5 & 85.2 & 95.9 & 85.7 & 90.0 & 83.5 & 89.5 & 96.9 & 89.1 & 95.4 & 85.0 & 93.6 & 85.2 & 91.6 \\
        TongUI~\citep{zhang2025tongui}
        & 7B & 91.9 & 79.5 & 93.8 & 80.0 & 89.1 & 81.6 & 86.0 & 93.1 & 81.5 & 96.4 & 82.9 & 90.2 & 84.7 & 88.7 \\
        GUI-Actor~\citep{wu2025gui}
        & 7B & 94.9 & 82.1 & 91.8 & 80.0 & 91.3 & 85.4 & 88.3 & 96.5 & 84.3 & 91.7 & 84.1 & 93.9 & 82.3 & 89.5 \\
        \midrule
        \multicolumn{16}{l}{\cellcolor{TableBlue}{\textit{GUI-specific Models (RL)}}} \\
        UI-R1~\citep{lu2025ui} 
        & 3B & 95.6 & 84.7 & 90.2 & 59.3 & 85.2 & 73.3 & 83.3 & 96.2 & 84.3 & 92.3 & 63.6 & 89.2 & 75.4 & 85.4 \\
        UI-R1-E~\citep{lu2025ui}
        & 3B & 97.1 & 83.0 & 95.4 & 77.9 & 91.7 & 85.0 & 89.2 & 98.2 & 83.9 & 94.8 & 75.0 & 83.7 & 93.2 & 89.5 \\ 
        SE-GUI~\citep{yuan2025enhancing}
        & 7B & - & - & - & - & - & - & 88.2 & - & - & - & - & - & - & 90.3 \\
        GUI-G$^2$\xspace~\citep{tang2025gui}
        & 7B & 96.7 & 90.8 & 95.9 & 88.6 & 90.9 & 86.9 & \textbf{92.0} & 98.3 & 91.9 & 95.4 & 89.3 & 94.0 & 87.7 & \underline{93.3} \\
        \midrule
        \multicolumn{16}{l}{\cellcolor{TableBlue}{\textit{Ours}}} \\
        \multirow{2}{*}{HyperClick} 
        & 3B & 96.7 & 83.9 & 92.8 & 80.7 & 88.7 & 83.5 & 88.5 & 98.6 & 86.3 & 95.4 & 90.6 & 82.2 & 84.7 & 90.6 \\
        & 7B & 95.6 & 91.7 & 93.8 & 82.9 & 92.2 & 88.4 & \underline{91.5} & 98.3 & 93.4 & 96.9 & 85.7 & 96.2 & 86.7 & \textbf{93.7} \\
        \bottomrule
    \end{tabular}
}
\label{tab: screenspot_v1_v2}
\end{table*}

\subsection{Implementation Details}
We implement HyperClick based on Qwen2.5-VL-3B-Instruct and Qwen2.5-VL-7B-Instruct. Model training is conducted within the VLM-R1~\citep{shen2025vlm} codebase. We train for one epoch on 16 NVIDIA H100 GPUs, using a cosine schedule decaying from 1e-6 to 0, a global batch size of 16, 8 generations per instance, and a KL constraint coefficient of $\beta=0.04$. To improve efficiency, we leverage FlashAttention-2~\citep{dao2023flashattention}, adopt bfloat16 precision, and enable gradient checkpointing. During inference, the temperature is fixed to 0 to ensure reproducibility. Full details of the training data, including source datasets and the RL sample construction procedure, are provided in the Appendix.

\begin{table*}[!htbp]
\centering
\caption{GUI grounding accuracy on the ScreenSpot-Pro~\citep{li2025screenspot} benchmark over the CAD, Development, Creative, Scientific, Office, and OS sub-tasks. \textbf{Bold} and \underline{underline} indicate the best and second-best results.}
\resizebox{\linewidth}{!}
{
    \begin{tabular}{lcccccccccccccc}
        \toprule
        \multirow{2}{*}{\textbf{Model}} & \multirow{2}{*}{\textbf{Size}} & \multicolumn{2}{c}{\textbf{CAD}} & \multicolumn{2}{c}{\textbf{Development}} & \multicolumn{2}{c}{\textbf{Creative}} & \multicolumn{2}{c}{\textbf{Scientific}} & \multicolumn{2}{c}{\textbf{Office}} & \multicolumn{2}{c}{\textbf{OS}} & \multirow{2}{*}{\textbf{Avg.}} \\
        \cmidrule(lr){3-4}
        \cmidrule(lr){5-6}
        \cmidrule(lr){7-8}
        \cmidrule(lr){9-10}
        \cmidrule(lr){11-12}
        \cmidrule(lr){13-14}
        & & Text & Icon & Text & Icon & Text & Icon & Text & Icon & Text & Icon & Text & Icon \\
        \midrule
        \multicolumn{15}{l}{\cellcolor{TableBlue}{\textit{General Models}}} \\
        GPT-4o~\citep{openai2024hello} 
        & - & 2.0 & 0.0 & 1.3 & 0.0 & 1.0 & 0.0 & 2.1 & 0.0 & 1.1 & 0.0 & 0.0 & 0.0 & 0.8 \\
        Claude~\citep{anthropic2024claude} 
        & - & 14.5 & 3.7 & 22.0 & 3.9 & 25.9 & 3.4 & 33.9 & 15.8 & 30.1 & 16.3 & 11.0 &  4.5 & 17.1 \\
        Qwen2.5-VL~\citep{bai2025qwen2}
        & 7B & 16.8 & 1.6 & 46.8 & 4.1 & 35.9 & 7.7 & 49.3 & 7.3 & 52.5 & 20.8 & 37.4 & 6.7 & 26.8 \\
        \midrule
        \multicolumn{15}{l}{\cellcolor{TableBlue}{\textit{GUI-specific Models (SFT)}}} \\
        CogAgent~\citep{hong2024cogagent}
        & 18B & 7.1 & 3.1 & 14.9 & 0.7 & 9.6 & 0.0 & 22.2 & 1.8 & 13.0 & 0.0 & 5.6 & 0.0 & 7.7 \\
        SeeClick~\citep{cheng2024seeclick} 
        & 9.6B & 2.5 & 0.0 & 0.6 & 0.0 & 1.0 & 0.0 & 3.5 & 0.0 & 1.1 & 0.0 & 2.8 & 0.0 & 1.1 \\
        ShowUI~\citep{lin2025showui}
        & 2B & 2.5 & 0.0 & 16.9 & 1.4 & 9.1 & 0.0 & 13.2 & 7.3 & 15.3 & 7.5 & 10.3 & 2.2 & 7.7 \\
        Aria-UI~\citep{yang2024aria}
        & 25.3B & 7.6 & 1.6 & 16.2 & 0.0 & 23.7 & 2.1 & 27.1 & 6.4 & 20.3 & 1.9 & 4.7 &  0.0 & 11.3\\
        UGround~\citep{gou2025uground}
        & 7B & 14.2 & 1.6 & 26.6 & 2.1 & 27.3 & 2.8 & 31.9 & 2.7 & 31.6 & 11.3 & 17.8 & 0.0 & 16.5 \\
        UGround-V1~\citep{gou2025uground}
        & 7B & 15.8 & 1.2 & 51.9 & 2.8 & 47.5 & 9.7 & 57.6 & 14.5 & 60.5 & 13.2 & 38.3 & 7.9 & 45.2 \\
        OS-Atlas~\citep{wu2024atlas}
        & 7B & 12.2 & 4.7 & 33.1 & 1.4 & 28.8 & 2.8 & 37.5 & 7.3 & 33.9 & 5.7 & 27.1 & 4.5 & 18.9 \\
        UI-TARS~\citep{qin2025ui}
        & 7B & 20.8 & 9.4 & 58.4 & 12.4 & 50.0 & 9.1 & 63.9 & 31.8 & 63.3 & 20.8 & 30.8 & 16.9 & 35.7 \\
        TongUI~\citep{zhang2025tongui}
        & 7B & 17.3 & 9.4 & 40.9 & 3.5 & 31.3 & 7.0 & 50.7 & 12.7 & 45.8 & 13.2 & 28.0 & 6.7 & 24.7 \\
        GUI-Actor~\citep{wu2025gui}
        & 7B & - & - & - & - & - & - & - & - & - & - & - & - & 40.7 \\
        JEDI~\citep{xie2025scaling}
        & 7B & 38.0 & 14.1 & 42.9 & 11.0 & 50.0 & 11.9 & 72.9 & 25.5 & 75.1 & 47.2 & 33.6 & 16.9 & 39.5 \\
        \midrule
        \multicolumn{15}{l}{\cellcolor{TableBlue}{\textit{GUI-specific Models (RL)}}} \\
        UI-R1~\citep{lu2025ui}
        & 3B & 11.2 & 6.3 & 22.7 & 4.1 & 27.3 & 3.5 & 42.4 & 11.8 & 32.2 & 11.3 & 13.1 & 4.5 & 17.8 \\
        UI-R1-E~\citep{lu2025ui}
        & 3B & 37.1 & 12.5 & 46.1 & 6.9 & 41.9 & 4.2 & 56.9 & 21.8 & 65.0 & 26.4 & 32.7 & 10.1 & 33.5 \\
        GUI-R1~\citep{luo2025gui}
        & 7B & 23.9 & 6.3 & 49.4 & 4.8 & 38.9 & 8.4 & 55.6 & 11.8 & 58.7 & 26.4 & 42.1 & 16.9 & 31.3 \\
        InfiGUI-R1~\citep{liu2025infigui}
        & 3B & 33.0 & 14.1 & 51.3 & 12.4 & 44.9 & 7.0 & 58.3 & 20.0 & 65.5 & 28.3 & 43.9 & 12.4 & 35.7 \\
        GUI-G1~\citep{zhou2025gui}
        & 3B & 39.6 & 9.4 & 50.7 & 10.3 & 36.6 & 11.9 & 61.8 & 30.0 & 67.2 & 32.1 & 23.5 & 10.6 & 37.1 \\
        SE-GUI~\citep{yuan2025enhancing}
        & 7B & 51.3 & 42.2 & 68.2 & 19.3 & 57.6 & 9.1 & 75.0 & 28.2 & 78.5 & 43.4 & 49.5 & 25.8 & 47.3 \\
        GUI-G$^2$\xspace~\citep{tang2025gui} 
        & 7B & 55.8 & 12.5 & 68.8 & 17.2 & 57.1 & 15.4 & 77.1 & 24.5 & 74.0 & 32.7 & 57.9 & 21.3 & \underline{47.5} \\
        \midrule
        \multicolumn{15}{l}{\cellcolor{TableBlue}{\textit{Ours}}} \\
        \multirow{2}{*}{HyperClick}
        & 3B & 43.7 & 23.5 & 62.4 & 20.0 & 50.5 & 12.6 & 55.6 & 30.0 & 63.9 & 37.8 & 41.1 & 20.2 & 41.3 \\
        & 7B & 51.3 & 20.3 & 70.2 & 22.1 & 57.6 & 20.3 & 76.4 & 30.9 & 70.1 & 30.2 & 56.1 & 22.5 & \textbf{48.2} \\
        \bottomrule
    \end{tabular}
}
\label{tab: screenspot_pro}
\end{table*}

\subsection{Evaluation Benchmarks}
We comprehensively evaluated HyperClick's GUI grounding capability across four benchmarks. \textbf{ScreenSpot}~\citep{cheng2024seeclick} evaluates the grounding of the GUI on mobile, desktop, and web platforms, providing a diverse set of interface types for comparing robustness across common user scenarios. \textbf{ScreenSpot-V2}~\citep{wu2024atlas} extends ScreenSpot with more challenging tasks and refined annotations, and tests grounding accuracy in various real-world environments. \textbf{ScreenSpot-Pro}~\citep{li2025screenspot} focuses on high-resolution professional settings with expert-annotated tasks, covering 23 applications, five industries, and three operating systems. \textbf{MMBench-GUI}~\citep{wang2025mmbench} organizes tasks into a hierarchical structure of basic and advanced instructions, enabling systematic evaluation across instruction complexity levels.

\subsection{Main Results}
\begin{table*}[htbp]
\centering
\caption{GUI grounding accuracy on the MMBench-GUI~\citep{wang2025mmbench} benchmark over the Windows, MacOS, Linux, iOS, Android, and Web sub-tasks. \textbf{Bold} and \underline{underline} indicate the best and second-best results.}
\resizebox{\linewidth}{!}
{
    \begin{tabular}{lcccccccccccccc}
        \toprule
        \multirow{2}{*}{\textbf{Model}} & \multirow{2}{*}{\textbf{Size}} & \multicolumn{2}{c}{\textbf{Windows}} & \multicolumn{2}{c}{\textbf{MacOS}} & \multicolumn{2}{c}{\textbf{Linux}} & \multicolumn{2}{c}{\textbf{iOS}} & \multicolumn{2}{c}{\textbf{Android}} & \multicolumn{2}{c}{\textbf{Web}} & \multirow{2}{*}{\textbf{Avg.}} \\
        \cmidrule(lr){3-4}
        \cmidrule(lr){5-6}
        \cmidrule(lr){7-8}
        \cmidrule(lr){9-10}
        \cmidrule(lr){11-12}
        \cmidrule(lr){13-14}
        & & Basic & Adv. & Basic & Adv.& Basic & Adv.& Basic & Adv.& Basic & Adv.& Basic & Adv. & \\
        \midrule
        \multicolumn{15}{l}{\cellcolor{TableBlue}{\textit{General Models}}} \\
        GPT-4o~\citep{openai2024hello} 
        & - & 1.5 & 1.1 & 8.7 & 4.3 & 1.1 & 1.0 & 5.1 & 3.3 & 2.5 & 1.4 & 3.2 & 2.9 & 2.9 \\
        Claude~\citep{anthropic2024claude}
        & - & 1.5 & 0.7 & 12.5 & 7.5 & 1.1 & 0.0 & 13.7 & 10.6 & 1.4 & 1.4 & 3.2 & 2.3 & 4.7 \\
        Qwen-Max-VL~\citep{baiqwenvl}
        & - & 43.9 & 36.8 & 58.8 & 56.1 & 53.9 & 30.1 & 77.4 & 59.1 & 79.5 & 70.1 & 74.8 & 58.8 &  58.0 \\
        Qwen2.5-VL~\citep{bai2025qwen2}
        & 7B & 31.4 & 16.5 & 31.3 & 22.0 & 21.5 & 12.2 & 66.6 & 55.2 & 35.1 & 35.2 & 40.3 & 32.5 & 33.9 \\
        InternVL3~\citep{zhu2025internvl3}
        & 72B & 70.1 & 42.6 & 75.7 & 52.3 & 59.2 & 41.3 & 93.6 & 80.6 & 92.7 & 78.6 & 90.7 & 65.9 &  72.2 \\
        \midrule
        \multicolumn{15}{l}{\cellcolor{TableBlue}{\textit{GUI-specific Models (SFT)}}} \\
        ShowUI~\citep{lin2025showui}
        & 2B & 9.2 & 4.4 & 24.1 & 10.4 & 25.1 & 11.7 & 29.0 & 19.7 & 17.4 & 8.7 & 22.9 & 12.7 & 16.0 \\
        OS-Atlas~\citep{wu2024atlas}
        & 7B & 36.9 & 18.8 & 44.4 & 21.7 & 31.4 & 13.3 & 74.8 & 48.8 & 69.6 & 46.8 & 61.3 & 35.4 & 41.4 \\
        Aguvis~\citep{xu2024aguvis}
        & 7B & 37.3 & 21.7 & 48.1 & 33.3 & 33.5 & 25.0 & 67.5 & 65.2 & 61.0 & 51.0 & 61.6 & 45.5 & 45.7 \\
        UGround-V1~\citep{gou2025uground}
        & 7B & 66.8 & 39.0 & 71.3 & 48.6 & 56.5 & 31.1 & 92.7 & 70.9 & 93.5 & 71.0 & 88.7 & 64.6 &  65.7 \\
        UI-TARS~\citep{qin2025ui}
        & 72B & 78.6 & 51.8 & 80.3 & 62.7 & 68.6 & 51.5 & 90.8 & 81.2 & 93.0 & 80.0 & 88.1 & 68.5 & \underline{74.3} \\
        \midrule
        \multicolumn{15}{l}{\cellcolor{TableBlue}{\textit{Ours}}} \\
        \multirow{2}{*}{HyperClick}
        & 3B & 73.8 & 45.6 & 80.3 & 52.9 & 66.5 & 35.7 & 91.4 & 72.7 & 92.4 & 74.9 & 89.1 & 60.1 & 71.4 \\
        & 7B & 82.3 & 61.4 & 82.9 & 67.1 & 66.5 & 48.0 & 94.0 & 82.1 & 95.8 & 85.1 & 93.2 & 85.1 & \textbf{79.6} \\
        \bottomrule
    \end{tabular}
}
\label{tab: mmbench_gui}
\end{table*}
\textbf{Comparisons with recent baselines.}
We present the main results of our evaluation in Table~\ref{tab: screenspot_v1_v2},~\ref{tab: screenspot_pro},~\ref{tab: mmbench_gui}. The results show that HyperClick maintains competitive performance among open-source models in both 3B and 7B parameter categories while adding explicit confidence-quality alignment. In particular, HyperClick demonstrates consistent gains across diverse platforms and task settings, including mobile, desktop, and web environments. The benefits are especially visible in challenging benchmarks, such as ScreenSpot-Pro and MMBench-GUI, which feature high-resolution interfaces, small target elements, and complex visual layouts. This suggests that HyperClick is effective in scenarios where precise spatial grounding and robustness are critical.

A key source of HyperClick's improvement lies in the introduction of SCRL, which trains the model to pair each click prediction with an explicit confidence estimate. Unlike prior RL-based GUI grounding models that rely solely on sparse binary~\citep{lu2025ui,luo2025gui} or continuous~\citep{yuan2025enhancing,tang2025gui} correctness rewards, HyperClick leverages confidence alignment to distinguish reliable clicks from uncertain ones. This enables policy models to penalize overconfident errors while reinforcing well-aligned predictions. The resulting training objective yields consistent gains across benchmarks, suggesting that confidence alignment can improve generalization across diverse UI environments. These results show that trustworthy GUI grounding can enhance accuracy while providing a spatially grounded ranking signal that correlates with grounding success.

\begin{figure*}[!t]
    \centering
    \includegraphics[width=0.99\linewidth]{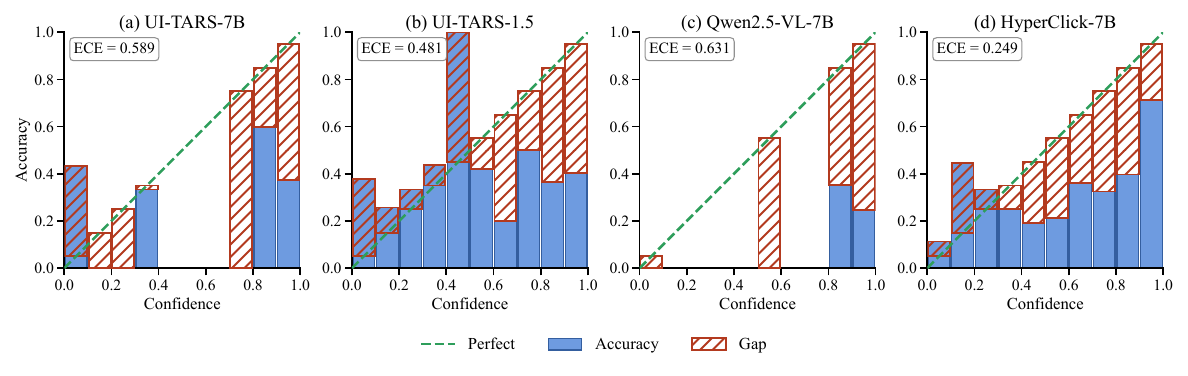}
    \caption{Reliability diagrams on ScreenSpot-Pro. Blue bars are the per-bin accuracy, red hatched bars are the gap to the perfect-alignment diagonal (green dashed), and the ECE value is shown in each panel. (a)~UI-TARS-7B, (b)~UI-TARS-1.5, and (c)~Qwen2.5-VL-7B are severely overconfident, whereas (d)~HyperClick-7B substantially reduces ECE and shows smaller alignment gaps across most confidence bins.}
    \label{fig: confidence}
\end{figure*}

\textbf{HyperClick improves confidence alignment.}
Figure~\ref{fig: confidence} reports reliability diagrams on ScreenSpot-Pro. For each confidence bin, the blue bar denotes empirical grounding accuracy, the red hatched region denotes the gap to the perfect-alignment diagonal, and the expected Calibration Error (ECE, $\downarrow$)~\citep{naeini2015obtaining,guo2017calibration} summarizes the sample-weighted alignment gap across bins.
The baseline models are severely overconfident, with large gaps in high-confidence bins and ECEs of 0.589, 0.481, and 0.631. In contrast, HyperClick-7B reduces ECE to 0.249 and yields visibly smaller alignment gaps across most bins, showing that its verbalized confidence better tracks empirical grounding accuracy.

\subsection{Ablation Study}
We conducted an ablation study on ScreenSpot-V2~\cite{cheng2024seeclick} to verify the key components of HyperClick.

\textbf{Reward Mechanism.}
The results in Table~\ref{tab: ablation_reward} demonstrate the need to combine format, correctness, and confidence alignment rewards. Using only the format reward yields relatively limited improvements (89.3\%), and using only the confidence alignment reward (2.1\%) causes reward hacking, where the policy model is optimized to predict only incorrect answers with confidence of 0. Moreover, we encourage the policy model to express confidence rounded to three decimal places and constrain it in $R_{\text{format}}$ to avoid overly long completions. This strict format verification is crucial for training stability and convergence, improving accuracy from 91.0\% to 93.7\%. In general, the combination of format, correctness, and confidence alignment rewards has the best performance. This validates our motivation that confidence alignment acts as an auxiliary uncertainty signal, discouraging overconfident errors and reinforcing reliable predictions.

\textbf{Alignment Formulation.}
We further examine the sensitivity of HyperClick to the specific confidence-alignment formulation in Table~\ref{tab: ablation_reward_formulation}. Following the principle that $R_{\text{confidence}}$ should be maximized when the predicted confidence $\hat{c}$ matches the spatial confidence target $\overline{\mathcal{C}}(\hat{x},\hat{y})$, we compare $l_1$, $l_2$, and KL variants. All three formulations improve over the no-confidence baseline, showing that the benefit does not depend on a single distance function. The $l_2$ formulation performs best (93.7\%) and is therefore used as the default, while KL achieves a comparable 93.6\%.

\begin{table}[!t]
\begin{minipage}{\linewidth}
    \centering
    \caption{Ablation study of reward configurations. * denotes reward hacking.}
    \label{tab: ablation_reward}
    \small
    \begin{tabular}{cccc}
        \toprule
        $R_{\text{format}}$
        & $R_{\text{correctness}}$
        & $R_{\text{confidence}}$
        & \textbf{Acc(\%)} \\
        \midrule
        {\checkmark} & & & 89.3 \\
        & \checkmark & & 92.3 \\
        & & \checkmark & $\text{2.1}^*$ \\
        \checkmark & \checkmark & & 92.3 \\
        & \checkmark & \checkmark & 91.0 \\
        \rowcolor{TableBlue} \checkmark & \checkmark  & \checkmark & 93.7 \\
        \bottomrule
    \end{tabular}
\end{minipage}

\vspace{0.6em}
\begin{minipage}{\linewidth}
    \centering
    \caption{Ablation study of confidence reward formulations. We denote $\bar{C}=\overline{\mathcal{C}}(\hat{x},\hat{y})$ and $\Delta=|\hat{c}-\bar{C}|$.}
    \label{tab: ablation_reward_formulation}
    \small
    \begin{tabular}{llc}
        \toprule
        \textbf{Type}
        & \textbf{Formulation}
        & \textbf{Acc(\%)} \\
        \midrule
        -- & -- & 92.3 \\
        $l_1$ & $1-\Delta$ & 93.2 \\
        \rowcolor{TableBlue}
        $l_2$ & $1-\Delta^2$ & 93.7 \\
        KL & $\exp\!\left(-\mathrm{KL}(\bar{C}\,\|\,\hat{c})\right)$ & 93.6 \\
        \bottomrule
    \end{tabular}
\end{minipage}

\vspace{0.6em}
\begin{minipage}[t]{0.45\linewidth}
    \centering
    \caption{Ablation study of confidence modeling.}
    \label{tab: ablation_confidence}
    \small
    \begin{tabular}{cc}
    \toprule
    \makebox[0.35\linewidth][c]{$\alpha$}
    & {\textbf{Acc(\%)}} \\
    \midrule
    0 & 92.7 \\
    1/2 &  93.1 \\
    \rowcolor{TableBlue} 1/4  & 93.7 \\
    1/6 & 92.1 \\
    \bottomrule
    \end{tabular}
\end{minipage}
\hfill
\begin{minipage}[t]{0.5\linewidth}
    \centering
    \caption{Ablation of baseline.}
    \label{tab: ablation_baseline}
    \small
    \begin{tabular}{lc}
    \toprule
    \textbf{Model} & \textbf{Acc(\%)} \\
    \midrule
    Qwen2.5-VL & 88.8  \\
    \rowcolor{TableBlue} HyperClick & 93.7 \\
    \midrule
    MiMo-VL & 91.4 \\
    \rowcolor{TableBlue} HyperClick & 94.0 \\
    \bottomrule
    \end{tabular}
\end{minipage}
\vspace{-1.2em}
\end{table}

\textbf{Confidence Modeling.}
Table~\ref{tab: ablation_confidence} investigates the effect of the adaptive variance factor $\alpha$. Without the bounded spatial confidence target ($\alpha$=0), only binary confidence is used for confidence alignment. Therefore, when $\alpha$=0, the confidence alignment reward is represented as:
\begin{equation}
    R_{\text{confidence}} = 1 - (\hat{c}-\mathbbm{1}_{\hat{\mathbf{p} \in \mathbf{b}}})^2.
\end{equation}
The policy model reaches 92.7\%, which is weaker than the Gaussian-target variants but still demonstrates the effectiveness of SCRL. We set $\alpha$ according to the Gaussian 3$\sigma$ principle: in the $x$ direction, $k\cdot\sigma_x=\frac{1}{2}(x_2-x_1)$, where $k \in \{1,2,3\}$ gives $\alpha \in \{\frac{1}{2}, \frac{1}{4}, \frac{1}{6}\}$. Both too large ($\alpha=\tfrac{1}{2}$) and too small ($\alpha=\tfrac{1}{6}$) variances are suboptimal: the former makes the target too diffuse across the box and weakens center-sensitive supervision, while the latter over-concentrates confidence near the center and becomes too strict for minor deviations. $\alpha=\tfrac{1}{4}$ provides the best concentration-spread trade-off and highest precision (93.7\%).

\textbf{Extension to other baselines.}
As shown in Table~\ref{tab: ablation_baseline}, we further extend HyperClick to MiMo-VL~\citep{coreteam2025mimovltechnicalreport}, a strong general-purpose VLM. With our training framework, MiMo-VL improves from 91.4\% to 94.0\%, demonstrating that HyperClick serves as a general training paradigm for GUI grounding, confirming the generality and scalability of our approach across various models.

\subsection{Visualization}
\begin{figure}[!t]
    \centering
    \includegraphics[width=0.99\linewidth]{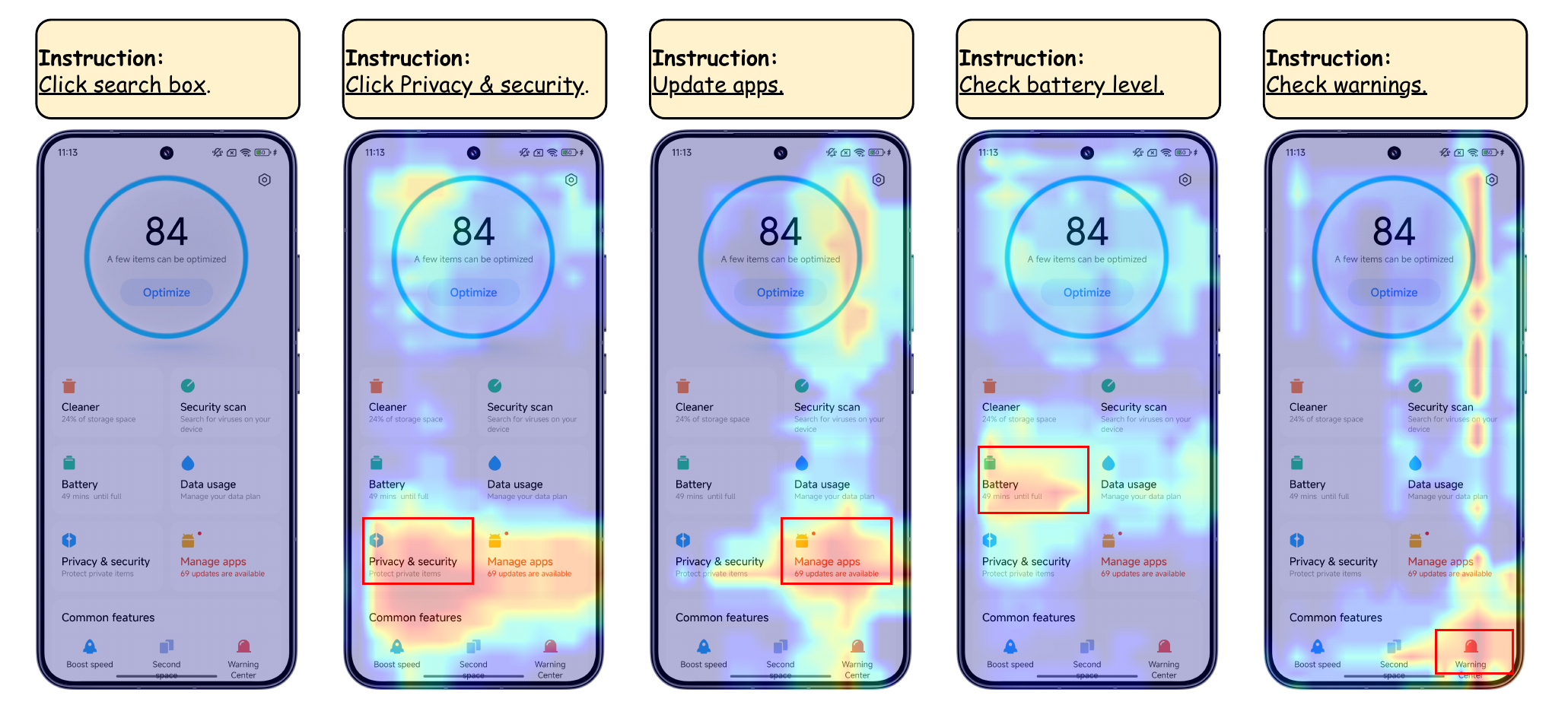}
    \caption{Visualization of the confidence distribution output by HyperClick. We inject the coordinates on the interface into the assistant's generation and enforce it to continue to output the confidence for the click position. The lighter the color, the higher the confidence value.}
    \label{fig: visualization}
\vspace{-1em}
\end{figure}

To better understand confidence alignment, Figure~\ref{fig: visualization} visualizes HyperClick's predicted confidence distributions. We inject interface coordinates into the assistant generation and require the policy to continue outputting confidence for each click position, yielding heatmaps over possible clicks. Confidence concentrates around ground-truth elements and stays low in irrelevant regions, matching the bounded spatial target design. Adaptive variance further adjusts the spread by element size: smaller UI elements yield tighter peaks, whereas larger ones produce broader heatmaps. 


\section{Conclusion}  
In this work, we address overconfidence in GUI grounding models, which undermines the reliability of autonomous GUI agents. We introduce HyperClick, a SCRL framework that augments grounding with explicit confidence alignment. By combining binary correctness reward with an $l_2$ confidence alignment reward derived from a bounded spatial confidence target, HyperClick improves grounding accuracy while producing better-aligned confidence estimates. Extensive experiments show strong accuracy and confidence alignment, suggesting a path toward trustworthy GUI grounding in broader multimodal agents.

\section*{Limitation}
Although the effect of the confidence alignment mechanism proposed in this work has been verified, it has not been extended to GUI planning tasks. We believe that the reliability of planning is even more critical for the overall success of GUI automation, since inaccurate or overconfident planning decisions can propagate errors across multiple steps and ultimately lead to task failure. In future work, we plan to investigate how confidence alignment can be incorporated into planning modules, enabling agents to not only ground actions reliably but also expose uncertainty during high-level decisions throughout complex multi-step interactions.

\section*{Ethics Consideration}
This research focuses on building a policy model for reliable GUI grounding. The data used are obtained by synthesizing or reprocessing previously released datasets, with all datasets or benchmarks properly cited. In this paper, there are no discrimination, bias, or fairness issues that need to be addressed. In addition, our models are not expected to generate potentially harmful content. To ensure reproducibility, we provide all experimental and data details in Section \ref{sec: experiments} and the corresponding appendices. We will release the source code and model checkpoints to support
reproducibility.

\bibliography{ref}

\clearpage
\appendix
\makeatletter
\setlength{\@fptop}{0pt}
\setlength{\@dblfptop}{0pt}
\makeatother

\section{Appendix}
\label{sec:appendix}

\subsection{Data Details}
\label{sec: data_details}
To provide a comprehensive grounding resource across diverse platforms, we construct a dataset containing 50K samples distributed across three representative domains: Mobile, Web, and Desktop. Training data is sampled from multiple public GUI datasets, including OS-Atlas~\citep{wu2024atlas}, Widget Caption~\citep{li2020widget}, UI-Refexp~\citep{bai2021uibert}, and OmniAct~\citep{kapoor2024omniact}, together with in-house data. Each domain contains a balanced set of grounding instances that pair natural language commands with corresponding UI elements. The per-source statistics are shown in Table~\ref{tab: data}.

\begin{table}[H]
    \centering
    \caption{Statistics and sources of the grounding dataset adopted in HyperClick.}
    \label{tab: data}
    \resizebox{\linewidth}{!}
    {
        \begin{tabular}{lccccccc}
            \toprule
            \textbf{Source} & OmniAct & ShowUI-Web & UI-Refexp & Widgnt-Caption & OS-Atlas & In-House\\
            \midrule
            \textbf{Size} & 119 & 19172 & 280 & 3672 & 26114 & 1664 \\
            \bottomrule
        \end{tabular}
    }
\end{table}

To construct high-quality samples for RL, we first employ Qwen2.5-VL-7B~\citep{bai2025qwen2} to generate raw data with the temperature set to 0, and identify cases where the model produces incorrect predictions. For each of these error cases, we then perform eight additional inferences with temperature 0.9 and extract the correctly predicted results as the final training data. In addition, prior to RL, we incorporate an equal number of correctly predicted samples from Stage 1 to provide a cold start. This initialization not only stabilizes the training but also helps the model adhere to the target output format: \texttt{<point>[x,y]</point>\allowbreak <confidence>conf</confidence>}, where \texttt{conf} is rounded to three decimal places.

\subsection{Evaluation Prompts}
\label{sec: prompts}
In this section, we detail the replicated evaluation prompts in ScreenSpot-Pro~\citep{li2025screenspot}. We follow the instructions they originally provided to reproduce and analyze the experimental results. The prompts are shown as follows:

\begin{tcolorbox}[colframe=PromptBlue, colback=TableBlue, coltitle=black!80, title=GPT-4o's Prompt]
Locate the UI element most related to the instruction \{\texttt{problem}\} on the screenshot.  Output only a JSON in the format [\{``point\_2d'': [...]\}].
\end{tcolorbox}

\begin{tcolorbox}[colframe=PromptBlue, colback=TableBlue, coltitle=black!80, title=Seed-VL's Prompt]
Locate the UI element most related to the instruction \{\texttt{problem}\} on the screenshot.  Output only a JSON in the format [\{``point\_2d'': [...]\}].
\end{tcolorbox}

\begin{tcolorbox}[colframe=PromptBlue, colback=TableBlue, coltitle=black!80, title=Qwen2.5-VL's Prompt]
Locate the UI element most related to the instruction \{\texttt{problem}\} on the screenshot.  Output only a JSON in the format [\{``point\_2d'': [...], ``label'': ... \}].
\end{tcolorbox}

\begin{tcolorbox}[colframe=PromptBlue, colback=TableBlue, coltitle=black!80, title=KiMi-VL's Prompt]
Point to the UI element most related to the instruction \{\texttt{problem}\} on the screenshot.
\end{tcolorbox}

\begin{tcolorbox}[colframe=PromptBlue, colback=TableBlue, coltitle=black!80, title=MiMo-VL's Prompt]
Locate the UI element most related to the instruction \{\texttt{problem}\} on the screenshot. Output a JSON format [\{``bbox\_2d'': [...], ``label'': ...\}]./no\_think
\end{tcolorbox}

\begin{tcolorbox}[colframe=PromptBlue, colback=TableBlue, coltitle=black!80, title=UI-TARS' and UI-TARS-1.5's Prompt]
Point to the element related to the instruction \{\texttt{problem}\} on the screenshot.
\end{tcolorbox}

Due to UI-TARS~\citep{qin2025ui} and UI-TARS-1.5~\citep{qin2025ui} being trained with a large amount of GUI-specific data, the ability to follow instructions is relatively poor. To prompt such models to generate verbalized confidence in their predictions, we adopt a multi-round conversation to output confidence for their answer. Specifically, policy models use the above prompts for GUI grounding in the first round and in the second round, generate the verbalized confidence of the prediction according to the prompt below:

\begin{tcolorbox}[colframe=PromptBlue, colback=TableBlue, coltitle=black!80, title=Confidence Prompt]
Output only a float number ranging from 0 to 1, representing your confidence with your provided answer, without any format.
\end{tcolorbox}

\subsection{Training Dynamics}
\label{sec: training_dynamics}
\begin{figure}[H]
    \centering
    \begin{subfigure}[t]{\linewidth}
        \centering
        \includegraphics[width=\linewidth]{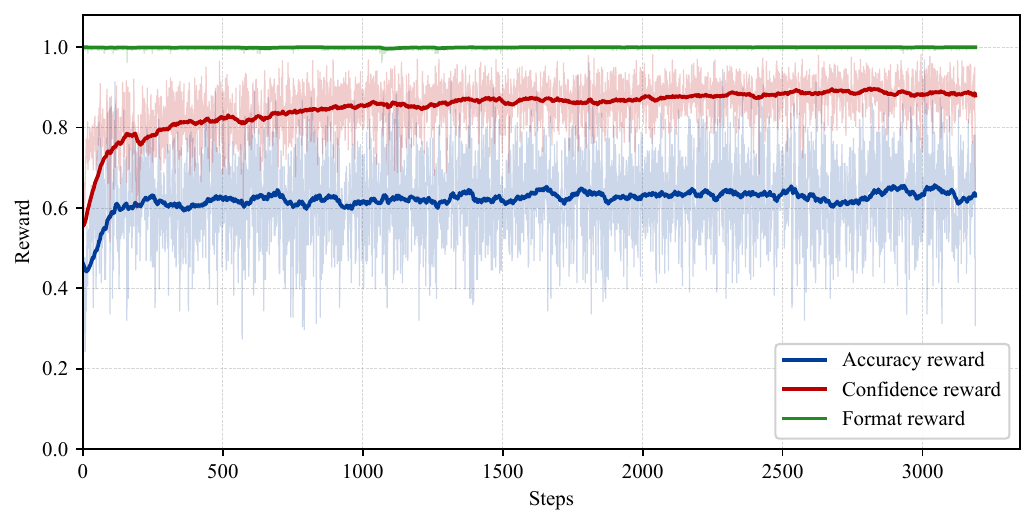}
        \caption{Mean reward over training steps.}
        \label{fig: rewards}
    \end{subfigure}
    \vspace{0.5em}
    \begin{subfigure}[t]{\linewidth}
        \centering
        \includegraphics[width=\linewidth]{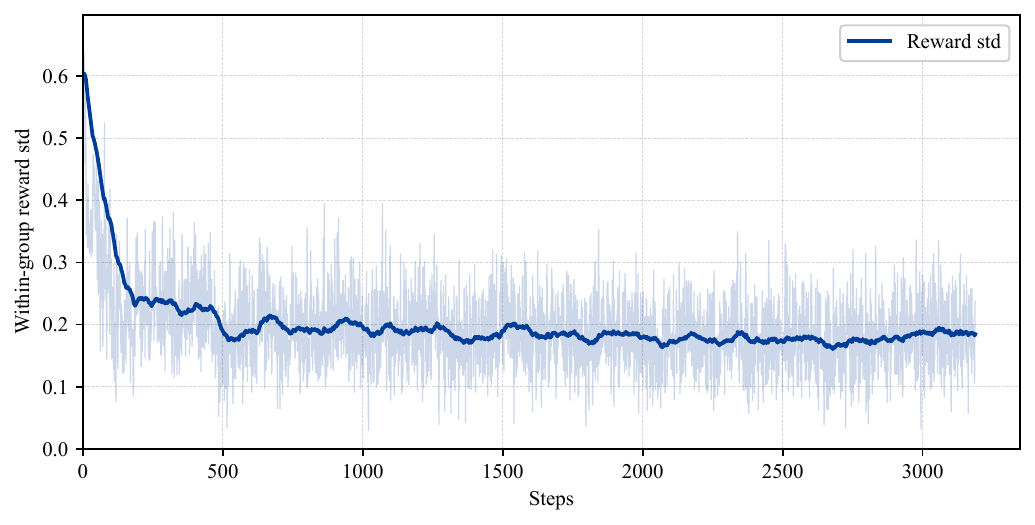}
        \caption{Reward standard deviation over training steps.}
        \label{fig: reward_std}
    \end{subfigure}
    \caption{\textbf{GRPO training dynamics for HyperClick-3B and HyperClick-7B.} (a) The mean reward rises and stabilizes, indicating that the policy increasingly satisfies the self-critique reward. (b) The per-group reward standard deviation stays non-trivial, showing that GRPO retains rollout diversity and a usable advantage signal throughout training.}
    \label{fig: training_dynamics}
\end{figure}

Figure~\ref{fig: training_dynamics} tracks two GRPO training signals over the course of RL fine-tuning for HyperClick-3B and HyperClick-7B: the mean reward and the per-group reward standard deviation. The mean reward rises steadily and plateaus, indicating that the policy progressively produces grounding outputs that satisfy the self-critique reward. The reward standard deviation stays well above zero throughout training, showing that GRPO continues to sample diverse rollouts within each prompt group and thus retains a usable advantage signal rather than collapsing to a degenerate policy. Together, the two curves confirm that the self-critiqued objective drives stable optimization without sacrificing exploration.

\subsection{Stability of Confidence}
\label{sec: stability}
To evaluate the stability of HyperClick's confidence, we verify whether the model gives similar confidence estimates for the same answer. We first let HyperClick predict the coordinates without sampling. Then, we inject the coordinates into the assistant's generation and instruct it to continue outputting confidence at a temperature of 1.0 for 8 times. As shown in Table~\ref{tab: variance}, we report the mean variance for different sample sizes. The results indicate that both HyperClick-3B and HyperClick-7B maintain low variance across different sampling scales, with the larger 7B model showing slightly more stable outputs. This suggests that the confidence estimation of HyperClick is stable under repeated sampling.

\begin{table}[H]
    \centering
    \caption{Stability evaluation of the model for the same prediction.}
    \label{tab: variance}
    \resizebox{\linewidth}{!}
    {
        \begin{tabular}{lcccccc}
            \toprule
            \multirow{2}{*}{\textbf{Model}} & \multicolumn{6}{c}{\textbf{Variance}} \\
            \cmidrule(lr){2-7}
            & 10 & 50 & 100 & 500 & 1000 & 1581 \\
            \midrule
            HyperClick-3B & 0.020 & 0.028 & 0.023 & 0.020 & 0.020 & 0.020 \\
            HyperClick-7B & 0.014 & 0.020 & 0.020 & 0.019 & 0.019 & 0.019 \\
            \bottomrule
        \end{tabular}
    }
\end{table}

\subsection{More Evaluation Benchmarks and Detailed Experimental Results}
\label{sec: more_benchmarks}
In this section, we present additional benchmarks and experimental results used in this work.

\textbf{CAGUI} is a Chinese benchmark for mobile GUI grounding. It emphasizes the grounding of textual elements and functional operations within Chinese-language applications. Detailed experimental results and comparisons with baselines are shown in Table~\ref{tab: cagui}.

\begin{table*}[!t]
\centering
\caption{GUI grounding accuracy on the CAGUI~\citep{zhang2025agentcpm} benchmark over the Fun2Point, Text2Point, and Bbox2Text sub-tasks. \textbf{Bold} and \underline{underline} indicate the best and second-best results.}
    \begin{tabular}{lcccc}
    \toprule
    \textbf{Model} & \textbf{Size} & \textbf{Fun2Point} & \textbf{Text2Point} & \textbf{Avg.} \\ 
    \midrule
    \multicolumn{5}{l}{\cellcolor{TableBlue}{\textit{General Models}}} \\
    GPT-4o~\citep{openai2024hello}
    & - & 22.1 & 19.9 & 21.0 \\
    Qwen2.5-VL~\citep{bai2025qwen2}
    & 7B & 59.8 & 59.3 & 59.6 \\
    InternVL2.5~\citep{chen2024expanding}
    & 8B & 17.2 & 24.2 & 20.7 \\
    \midrule
    \multicolumn{5}{l}{\cellcolor{TableBlue}{\textit{GUI-specific Models (SFT)}}} \\
    OS-Genesis~\citep{sun2024genesis}
    & 7B & 8.3 & 5.8 & 7.1 \\
    OS-Atlas~\citep{wu2024atlas} 
    & 7B & 53.6 & 60.7 & 57.2 \\
    Aguvis~\citep{xu2024aguvis}
    & 7B & 60.8 & 76.5 & 68.7 \\
    UI-TARS~\citep{qin2025ui}
    & 7B & 56.8 & 66.7 & 61.8 \\
    \midrule
    \multicolumn{5}{l}{\cellcolor{TableBlue}{\textit{GUI-specific Models (RL)}}} \\
    AgentCPM-GUI~\citep{zhang2025agentcpm}
    & 8B & 79.1 & 76.5 & 77.8 \\
    \midrule
    \multicolumn{5}{l}{\cellcolor{TableBlue}{Ours}} \\
    \multirow{2}{*}{HyperClick}
    & 3B & 80.9 & 81.2 & \underline{81.0} \\
    & 7B & 82.7 & 83.1 & \textbf{82.9} \\
    \bottomrule
    \end{tabular}
\label{tab: cagui}
\end{table*}

\textbf{UI-I2E-Bench} introduces implicit instructions that require both semantic understanding and spatial reasoning. Highlights the limitations of direct grounding and encourages models to adopt more sophisticated reasoning. Detailed experimental results and comparisons with baselines are shown in Table~\ref{tab: ui_i2e}.

\begin{table*}[!t]
\centering
\caption{GUI grounding accuracy on the UI-I2E-Bench~\citep{liu2025ui} benchmark over the platforms of mobile, desktop, and web with various implicitness. \textbf{Bold} and \underline{underline} indicate the best and second-best results.}
{
    \begin{tabular}{lccccccc}
        \toprule
        \multirow{2}{*}{\textbf{Model}} & \multirow{2}{*}{\textbf{Size}} & \multicolumn{3}{c}{\textbf{Platform}} & \multicolumn{2}{c}{\textbf{Implicitness}} & \multirow{2}{*}{\textbf{Avg.}} \\
        \cmidrule(lr){3-5}
        \cmidrule(lr){6-7}
        & & Mobile & Desktop & Web & Explicit & Implicit \\
        \midrule
        \multicolumn{8}{l}{\cellcolor{TableBlue}{\textit{General Models}}} \\
        Qwen2.5-VL~\citep{bai2025qwen2}
        & 7B & 61.7 & 41.6 & 56.9 & 58.4 & 51.0 & 53.8 \\
        \midrule
        \multicolumn{8}{l}{\cellcolor{TableBlue}{\textit{GUI-specific Models (SFT)}}} \\
        ShowUI~\citep{lin2025showui}
        & 2B & 53.9 & 30.4 & 29.6 & 51.3 & 35.6 & 41.5 \\
        SeeClick~\citep{cheng2024seeclick}
        & 9.6B & 37.2 & 15.8 & 18.2 & 37.1 & 19.9 & 26.4 \\
        Aguvis~\citep{xu2024aguvis}
        & 7B & 60.3 & 47.6 & 45.1 & 61.1 & 48.4 & 53.2 \\
        OmniParser~\citep{wan2024omniparser}
        & - & 67.6 & 45.5 & 30.8 & 54.3 & 52.4 & 53.1 \\
        OmniParser~\citep{yu2025omniparser}
        & - & 69.4 & 42.4 & 40.7 & 57.0 & 53.5 & 54.8 \\
        OS-Atlas~\citep{wu2024atlas}
        & 7B & 68.1 & 48.9 & 52.2 & 63.2 & 55.8 & 58.6 \\
        UGround-V1~\citep{gou2025uground}
        & 7B & 73.5 & 65.7 & 70.8 & 81.3 & 63.6 & 70.3 \\
        UI-TARS~\citep{qin2025ui}
        & 7B & 65.7 & 58.0 & 56.5 & 71.4 & 55.3 & 61.4 \\
        UI-I2E-VLM~\citep{liu2025ui}
        & 7B & 76.2 & 64.0 & 62.1 & 72.0 & 67.9 & 69.5 \\
        \midrule
        \multicolumn{8}{l}{\cellcolor{TableBlue}{\textit{GUI-specific Models (RL)}}} \\
        UI-R1~\citep{lu2025ui}
        & 3B & 67.8 & 46.2 & 58.1 & 67.9 & 52.8 & 58.5\\
        \midrule
        \multicolumn{8}{l}{\cellcolor{TableBlue}{\textit{Ours}}} \\
        \multirow{2}{*}{HyperClick}
        & 3B & 77.9 & 59.0 &81.0 & 81.1 & 66.1 & \underline{71.8} \\
        & 7B & 80.4 & 67.5 & 84.2 & 84.8 & 71.4 & \textbf{76.5} \\
        \bottomrule
    \end{tabular}
}
\label{tab: ui_i2e}
\end{table*}

\textbf{UI-Vision} evaluates the generalization of cross-applications in diverse desktop environments. By incorporating previously unseen applications, it tests the model's robustness and adaptability. Detailed experimental results and comparisons with baselines are shown in Table~\ref{tab: ui_vision}.

\begin{table*}[!t]
\centering
\caption{GUI grounding accuracy on the UI-Vision~\citep{nayak2025ui} benchmark over the Education (Ed.), Browsers (Br.), Development (De.), Productivity (Pr.), Creativity (Cr.), and Entertainment (En.) subtasks. \textbf{Bold} and \underline{underline} indicate the best and second-best results.}
\resizebox{\linewidth}{!}
{
    \begin{tabular}{lccccccccccc}
        \toprule
        \multirow{2}{*}{\textbf{Model}} & \multirow{2}{*}{\textbf{Size}} & \multicolumn{3}{c}{\textbf{Setting}} & \multicolumn{6}{c}{\textbf{Category}} & \multirow{2}{*}{\textbf{Avg.}} \\
        \cmidrule(lr){3-5}
        \cmidrule(lr){6-11}
        & & Basic & Functional & Spatial & Ed. & Br. & De. & Pr. & Cr. & En. \\
        \midrule
        \multicolumn{12}{l}{\cellcolor{TableBlue}{\textit{General Models}}} \\
        GPT-4o~\citep{openai2024hello}
        & - & 1.6 & 1.5 & 1.0 & 1.5 & 0.0 & 2.2 & 1.1 & 0.8 & 4.2 & 1.4 \\
        Gemini-1.5-pro~\citep{team2024gemini} 
        & - & 0.8 & 0.3 & 0.6 & 0.5 & 0.6 & 0.9 & 0.5 & 0.4 & 0.0 & 0.6 \\
        Claude~\citep{anthropic2024claude}
        & - & 9.5 & 7.7 & 7.6 & 6.1 & 9.8 & 8.0 & 9.4 & 7.7 &  8.3 & 8.3 \\
        Qwen2.5-VL~\citep{wang2024qwen2}
        & 7B & 1.2 & 0.8 & 0.5 & 0.5 & 0.0 & 1.2 & 0.9 & 0.5 & 1.0 & 0.9\\
        InternVL2.5~\citep{chen2024expanding}
        & 8B & 2.5 & 2.8 & 1.0 & 1.1 & 7.0 & 3.0 & 1.8 & 1.2 & 5.2 & 2.1 \\
        MiniCPM-V~\citep{yao2024minicpm}
        & 8B & 7.1 & 5.3 & 1.5 & 3.0 & 16.8 & 5.4 & 3.8 & 2.1 & 13.0 & 4.3 \\
        \midrule
        \multicolumn{12}{l}{\cellcolor{TableBlue}{\textit{GUI-specific Models (SFT)}}} \\
        CogAgent~\citep{hong2024cogagent} 
        & 9B & 12.0 & 12.2 & 2.6 & 8.7 & 11.2 & 8.6 & 10.3 & 5.6 & 15.6 & 8.9 \\
        SeeClick~\citep{cheng2024seeclick}
        & 9.6B & 9.4 & 4.7 & 2.1 & 4.2 & 13.3 & 7.3 & 4.3 & 4.0 & 11.0 & 5.4 \\
        AriaUI~\citep{yang2024aria}
        & 25.3B & 12.2 & 14.0 & 4.0 & 9.0 & 18.9 & 11.2 & 10.4 & 6.5 & 19.3 & 10.1 \\
        ShowUI~\citep{lin2025showui}
        & 2B & 8.1 & 7.7 & 2.1 & 3.7 & 13.3 & 7.5 & 6.5 & 2.5 & 15.6 & 5.9 \\
        OS-Atlas~\citep{wu2024atlas}
        & 7B & 12.2 & 11.2 & 3.7 & 8.7 & 16.8 & 10.3 & 9.2 & 5.6 & 16.2 & 9.0 \\
        UGround-V1~\citep{nayak2025ui}
        & 7B & 15.4 & 17.1 & 6.3 & 10.4 & 28.7 & 17.5 & 12.2 & 8.6 & 18.2 & 12.9 \\
        Aguvis~\citep{xu2024aguvis}
        & 7B & 17.8 & 18.3 & 5.1 & 13.1 & 30.8 & 17.1 & 12.1 & 9.6 &  24.0 & 13.7 \\
        UI-TARS~\citep{qin2025ui}
        & 7B & 20.1 & 24.3  & 8.4 & 14.2 & 35.0 & 19.7 & 18.3 & 11.1 & 38.5 & 17.6 \\
        \midrule
        \multicolumn{12}{l}{\cellcolor{TableBlue}{\textit{Ours}}} \\
        \multirow{2}{*}{HyperClick} 
        & 3B & 28.7 & 24.4 & 6.8 & 19.6 & 30.8 & 20.6 & 21.1 & 12.7 & 40.6 & 19.6 \\
        & 7B & 35.3 & 32.1 & 11.0 & 24.3 & 47.6 & 26.5 & 27.1 & 18.3 & 50.0 & \textbf{25.7} \\
        \bottomrule
    \end{tabular}

}
\label{tab: ui_vision}
\end{table*}

\end{document}